\pdfoutput=1

\documentclass[11pt]{article}

\usepackage{ACL2023}

\usepackage{times}
\usepackage{latexsym}

\usepackage[T1]{fontenc}

\usepackage[utf8]{inputenc}

\usepackage{microtype}

\usepackage{inconsolata}

\usepackage{booktabs} 
\usepackage{multirow}
\usepackage{xcolor} 
\usepackage{color,soul}
\usepackage{enumitem}
\usepackage{graphicx}
\usepackage{mathtools}
\usepackage{nicefrac} 
\usepackage{siunitx}
\usepackage{caption}
\usepackage{comment}

\title{Estimating Agreement by Chance \\
for Sequence Annotation}

\author{Diya Li \\
  Freenome Holdings, Inc\\
  \texttt{916lidiya@gmail.com} \\\And
  Carolyn Rosé \\
Carnegie Mellon University \\
  \texttt{cprose@cs.cmu.edu} \\\AND
  Ao Yuan \\
  Georgetown University \\
  \texttt{ay312@georgetown.edu} \\\And
  Chunxiao Zhou \\
  National Institutes of Health \\
  \texttt{chunxiao.zhou@nih.gov} \\}

\begin{document}
\maketitle

\begin{abstract}
In the field of natural language processing, correction of performance assessment for chance agreement plays a crucial role in evaluating the reliability of annotations. However, there is a notable dearth of research focusing on chance correction for assessing the reliability of sequence annotation tasks, despite their widespread prevalence in the field. 
To address this gap, this paper introduces a novel model for generating random annotations, which serves as the foundation for estimating chance agreement in sequence annotation tasks. 
Utilizing the proposed randomization model and a related comparison approach, we successfully derive the analytical form of the distribution, enabling the computation of the probable location of each annotated text segment and subsequent chance agreement estimation. Through a combination simulation and corpus-based evaluation, we successfully assess its applicability and validate its accuracy and efficacy.
\end{abstract}


\section{Introduction}
Reliable annotation is a cornerstone of NLP research, enabling both supervised learning methods and evaluation. Though not frequently employed for evaluation of model performance in the field of NLP, one of the most widely accepted metrics for evaluation of annotation reliability is Cohen's Kappa, which offers an assessment of inter-rater reliability that is adjusted in order to avoid offering credit for the portion of observed agreement that can be attributed to chance. Some NLP tasks, such as Named Entity Recognition or other span detection/labeling tasks, lack an appropriate chance corrected metric. This paper addresses this gap by proposing such a measure for these tasks, demonstrating its application in both simulation and CoNLL03 corpus experiments.

Numerous studies caution against using non-chance-corrected agreement metrics. They can lead to unfair task or system comparisons due to biases introduced due to varying levels of chance agreement across tasks and systems \citep{1_ide2017handbook, 2_komagata2002chance, 3_gates2017impact, 4_rand1971objective, 5_lavelli2008evaluation, 6_artstein2008inter}. Furthermore, without correction for chance agreement, measurements tend to cluster within a narrow range, making it difficult to discern differences between approaches \citep{7_eugenio2004kappa}. Therefore, both estimating and correcting for chance agreement have become critical in annotation evaluation, except in cases where chance agreement is negligible.

The main contributions of our work are summarized as follows:
\begin{itemize}[leftmargin=*,noitemsep,topsep=4pt]
 \item We propose a novel random annotation model that considers the specific characteristics of sequence annotation tasks as well as the annotation tendencies of different annotators.  This model can be divided into sub-models, enabling us to separately address cases with or without annotation overlap.
 We also apply chance agreement to measure task difficulty.
 \item Due to the additive nature of many popular similarity measures, we simplify the modeling of dependent annotation segments within a text. We successfully derive analytical probability distributions for random annotations, presenting a streamlined formulation that avoids redundant calculations.
 \item We delve into the asymptotic properties of agreement by chance, highlighting scenarios where it can be disregarded.
 \item We design and implement both simulation-based and naturalistic experiments, demonstrating that our proposed method is accurate, effective, and computationally efficient.
\end{itemize}
In the remainder of the paper, we provide a theoretical foundation for our work through a review of past literature. We then explain our methodology, and evaluate it first through a simulation study, and then through application to real-world corpora. Finally, we conclude with discussions of limitations, ethical considerations, and future research.

\section{Theoretical Foundation and Motivation}
Estimation of chance agreement is a key element in the evaluation of classification tasks. However, though the field of NLP features a wide variety of span detection and labeling tasks, there is a lack of widely adopted chance-corrected metrics for them.

In classification tasks, the Kappa coefficient is one of the most popular chance-corrected inter-annotator agreement measures \citep{2_komagata2002chance, 6_artstein2008inter, 7_eugenio2004kappa, 8_hripcsak2005agreement, 9_powers2015f}. 
The Kappa coefficient is defined as $(A_o-A_e)/(1-A_e)$, where $A_o$ is the observed agreement without chance agreement correction, and $A_e$ is the expected agreement assuming random annotation behavior.  
To estimate the chance agreement $A_e$, the key problem is how to build a random annotation model with reasonable assumptions. 

\begin{table*}[t]
\begin{center}
\begin{small}
\footnotesize
    \begin{tabular}{p{1.48cm}|p{4.2cm}|p{4.2cm}|p{4.2cm}}
	 \toprule
	  & Observed & Random & Invalid Random \\
	 \midrule
    Annotator 1 & I visited \hl{the NIH campus }in \hl{MD} & 
    I \hl{visited} the \hl{NIH campus in} MD & 
    \hl{I} visited \hl{the NIH} campus \hl{in} MD \\
    Annotator 2 & I visited \hl{the NIH campus in MD} & 
    \hl{I visited the NIH campus} in MD & 
    I \hl{visited} the \hl{NIH campus in MD} \\
	 \bottomrule
	\end{tabular}
\caption{\label{tab_example}
Example of a Toy Named Entity Annotation. Highlighted texts are annotations.}
\end{small}
\end{center}
\end{table*} 

Chance-corrected agreement is unarguably desirable for the evaluation of complex text annotation tasks beyond classification. These tasks encompass sequence annotation tasks \citep{11_lampert2016empirical, 12_esuli2010evaluating, 14_dai2018recognizing}, which involve a wide array of challenges. The complexity arises from the fact that estimating chance agreement is notably more intricate in comparison to straightforward classification tasks. In classification, the decisions to be made and the available options for each decision are uniform among annotators. However, with span prediction tasks, annotators initially identify the spans requiring labeling and subsequently assign a category to each of these spans. Discrepancies can arise at either of these stages, resulting from variations in span selection or category assignment.

Let's consider the Named Entity Recognition (NER) task as an illustrative example. It's important to note that the quantity and size of recognized entities can significantly differ among various annotators working on the same text. In Table \ref{tab_example}, we provide an example of a simplified NER task with annotations from two annotators. The text comprises seven tokens, each represented by a single word. The "Observed" column in the table showcases the annotations made by these two annotators.
In this toy example, annotator 1 identified and labeled two location entities: "the NIH campus" consisting of 3 tokens, and "MD" with 1 tokens. Meanwhile, annotator 2 identified a single entity, "the NIH campus in MD" encompassing 5 tokens.

While estimating inter-annotator agreement has become a crucial step in annotation evaluation, the challenge of estimating chance agreement for sequence annotation remains an open problem. As highlighted by numerous prior studies, the sample space for a sequence annotation task is often not well-defined \citep{10_cunningham2014gate}.

For instance, when considering the variability in annotator preferences, some tend to combine adjacent information, while others prefer to label them as distinct spans. Additionally, some annotators choose to encompass surrounding text within a segment, whereas others aim for shorter spans. All of these factors contribute to the complexity of estimating chance agreement in the context of sequence annotation tasks.

There is very little research on estimating chance agreement for span prediction tasks like NER. To the best of our knowledge, the most comprehensive and in-depth attempts so far have been the family of Krippendorff's Alpha coefficients. Unlike Kappa, the Alpha coefficient is grounded in the concept of disagreement, represented as $1 - D_o/D_e$, where $D_o$ stands for observed disagreement, and $D_e$ denotes expected disagreement. 

In 1995, Krippendorff first attempted to extend his Alpha coefficient for classification tasks to sequence labeling tasks \citep{krippendorff1995reliability}. The approach involved concatenating all annotations by different annotators for the same text and generating two copies. One copy remained unaltered, while the other undergoes all possible cyclic shifts. Krippendorff estimated the expected disagreement by comparing the differences between pairs of segments across these two sets of annotations. However, this shift-based random annotation model lacks a solid theoretical foundation and exhibits sensitivity to the location of relevant segments.

In 2016, Krippendorff introduced another data-driven approach to estimate expected disagreement \citep{krippendorff2016reliability}. This technique compares the dissimilarities between pairs of segments annotated by different annotators. It heavily relies on a large-scale annotation dataset. Notably, as it combines all annotation data from diverse texts indiscriminately, it cannot differentiate between different chance agreements corresponding to different annotation tasks.

In addition, Mathet proposed the gamma coefficient as a new metric for sequence labeling in 2015.  The gamma coefficient paper \citep{mathet2015unified} extensively discusses the various applications and characteristics of sequence labeling tasks. Although the gamma coefficient has many contributions, such as combining an optimization of alignment in the computation of the measure, its estimation of expected chance agreement is in line with Krippendorff's work and differs fundamentally from our approach.


It is critical to emphasize that neither of Krippendorff's methods are suitable for sequence annotation tasks, especially within the context of information extraction. When calculating disagreement, the Alpha coefficient accounts for all disagreements between segment pairs, encompassing both relevant and irrelevant segments. In cases where relevant information is sparse, the Alpha coefficient may be disproportionately influenced by disagreements related to irrelevant information, regardless of the consistency of annotations for relevant content. However, in information extraction tasks, our primary concern typically focuses on the consistency of annotations related to portions of text with a high concentration of relevant information. In the experiments section, we will probe further into this issue by exploring the limitations of Alpha coefficients within the context of information extraction.
 
While the specific problem of estimating chance agreement for span prediction tasks is an open problem, we must acknowledge that some relevant research has been done in connection with classification and clustering problems that informs our work and provides a continuum that our work extends \citep{15_hennig2015handbook, 16_franti2014centroid, 17_rezaei2016set, 18_van2019understanding, 19_warrens2019understanding, 20_meilua2007comparing, 21_vinh2010information}. 
Estimating agreement by chance is relatively simple in classification, because the sample space is fixed and the same for each annotator. 

In contrast, clustering problems present a greater challenge and bear closer resemblance to span prediction issues. From a conceptual standpoint, one could draw a parallel between elements within the same span and elements within the same cluster. The most commonly employed randomization model in clustering is the permutation model \citep{3_gates2017impact}, where all potential clusters, each with a fixed number of clusters and a fixed cluster size, are randomly generated with equal probability. However, what distinguishes span prediction from clustering is that the permutation model in clustering doesn't impose any restrictions on the placement of elements within the same cluster. Elements within the same cluster can be positioned anywhere. This assumption isn't suitable for sequence annotations, where segments are most typically comprise contiguous elements rather than fragmented. In essence, annotators treat each segment as a whole, rather than labeling each token independently. 

The variation in sample spaces caused by different labeling tendencies and connectivity constraints within each segment makes this problem quite challenging, especially when annotated segments need to be non-overlapping. Therefore, considering the characteristics of span prediction tasks and different annotation tendencies, we propose a new random annotation model to fulfill these requirements.

Our random annotation model independently models each annotator's tasks. Specifically, given the observed annotations for each task by each annotator, our random model uniformly randomizes entity positions while preserving the respective number of entities and the length of each entity.

To cater to various application requirements, we have designed two sub-models: the overlapping model and the non-overlapping model. These sub-models can accommodate situations where tasks necessitate non-overlapping spans and situations where no such requirement is specified.

For example, in Table \ref{tab_example}, the "Random" column presents a sample of random annotations for each annotator. For annotator 1, the random annotation still consists of two entities: "NIH campus in" with 3 tokens and "visited" with 1 tokens, both with randomized positions. In contrast, the "Invalid random" column in Table \ref{tab_example} provides examples of invalid random annotations, as neither the number nor the length of entities matches the observed annotation. It's important to note that in the random annotation model, the number of entities and the length of each entity are fixed for each annotator for each task, but these may vary between annotators for the same task. This flexibility is a deliberate choice in the random annotation model to account for the distinct annotation tendencies of each annotator, resulting in different chance agreements.

As another motivating observation, we recognize that many similarity measures are additive. In essence, the comparison between the annotations of different annotators involves accumulating comparisons among all segment pairs annotated by different annotators. For example, one of the most popular metrics, the F1 score for binary classification, can be expressed as $2a/(2a+b+c)$, where $a$ represents the number of items labeled as positive by both annotators, and $b$ and $c$ indicate the numbers of items rated as positive by one annotator but negative by the other. It's important to note that when the number and length of spans are both observed, the value of $2a+b+c$ is a constant. The "positive agreement" rating, denoted as $a$, reflects the cumulative sum of positive agreements for all compared segment pairs.

To simplify the modeling of random sequence annotations, we approach each segment individually, even though each labeled segment is still influenced by constraints imposed by other labeled segments within the same text, particularly in situations where segment overlap is not allowed. We have successfully derived the analytical distribution for the location of each individually labeled segment. Additionally, we've observed that the probability remains relatively consistent across most segment locations, reducing the need for numerous redundant calculations. Further details will be presented in the next section.

\section{Method}
In this section, we provide the specification of the random annotation model for sequence annotation, also known as span prediction, and present the calculation, approximation, and asymptotic properties of chance agreement through random annotation.

Taking NER as an example, we begin by introducing random sequence annotation models for both non-overlapping and overlapping scenarios, accompanied by the mathematical definition of chance estimation. Leveraging additive similarity measures, we significantly simplify the estimation of expected chance agreement in \textit{Proposition 1}, alongside its corresponding analytical formula for the distribution of random annotations in \textit{Proposition 2}. In \textit{Proposition 3}, we emphasize that each randomly annotated segment exhibits the same probability for most locations, with the exception of a few at the extreme ends, thus further reducing computational complexity.

Moreover, for lengthy texts with sparse annotation information, the expected chance agreement becomes so negligible that it can be safely disregarded. This assertion is substantiated in \textit{Proposition 4}. The preceding conclusions primarily pertain to non-overlapping scenarios, and we briefly encapsulate the outcome for the overlapping model in \textit{Proposition 5}, as its derivation is straightforward. Given space constraints, we present only the primary conclusions and concepts within this section. For detailed proofs, please consult the appendix.

We adopt the NER as a representative of complex text sequence annotation tasks to demonstrate how to estimate the chance agreement or performance for sequence annotation evaluation. 
Given a text $T=\{t_1 \prec t_2 \prec \ldots \prec t_n\}$ with a sequence of $n$ tokens $t_i, i\in \{1, \dots, n\}$, and a pre-defined tag set $C=\{c_1, \ldots, c_m\}$ with $m$ categorical tags; as a typical task in information extraction, named entity recognition aims to locate and classify segments of text $T$ into pre-defined categories $C$, such as recognizing disease, medication, and symptom information from clinical notes. 

Mathematically, the annotation task for NER can be formulated as a function $\Phi: T \times C \mapsto \Omega$, where $\Omega$ is the set of all possible annotations. 
For any $\psi \in \Omega$, $\psi=\{\psi_{1,1}, \ldots, \psi_{1,k_1}, \ldots, \psi_{m,1}, \ldots, \psi_{m,k_m}\}$, where $\psi$ is an annotation of segments for all pre-defined categories, $k_i$ is the number of segments for $i$-th category. 
For an annotation segment $\psi_{i,j}=\{st_{i,j}, a_{i,j}\}$,  $st_{i,j}$ denotes the index of the first token and $a_{i,j}$ denotes the length for the $j$-th segment with $i$-th category. 
To simplify the discussion, in the following we will focus on single-tag text annotation (i.e., $m=1$, $\psi=\{\psi_1, \ldots,  \psi_k\}$, $\psi_{j}=\{st_{j}, a_{j}\}$) since it is straightforward to generalize these techniques to multi-tag annotation as shown in the experiments. 

To gauge chance agreement, we need a precise definition of random annotation. Adapting the permutation model, which is commonly used for clustering, to sequence annotation tasks is impractical due to the absence of location constraints within clusters. This conflicts with the usual intra-segment connectivity assumption in a text annotation setting. To overcome this, we propose a novel random annotation model. It accommodates annotator and task variation while upholding the coherence of text segments.

{\bf \textit{Random Sequence Annotation Model}}.
The random annotation model is designed to keep the count and length of annotated segments consistent for each annotator within each task, while allowing variability across different annotators and tasks. It generates all feasible annotation configurations with equal probability.
In other words, for a $k$-segment random annotation $\Psi=\{\Psi_{1},  \ldots, \Psi_{k}\}$ with each randomly annotated segment $\Psi_{i}=\{ST_{i}, a_{i}\}$,
it has equal probabilities for all possible start indices $\{st_{1}, \ldots, st_k\}$ with fixed lengths $a_{1}, \ldots, a_k$. 

For annotator 1 in Table 1, we have $k=2$, $a_{1}=3$, $ST_{1} \in \{1,\ldots,5\}$, and $a_{2}=1$, $ST_{2} \in \{1,\ldots,7\}$.
The definition of a random annotation segment $\{ST_{i}, a_{i}\}$ indicates its connectivity.
All tokens in the same segment are consecutive without gaps and the index of the last token in the $i$-th annotated segment is $ST_{i}+a_i-1$. 
In contrast, a random cluster generated by the permutation model for clustering does not require this property. 
Note that the permutation of different entities is still allowed in our model as long as the segments within each entity remain contiguous, in other words, that the entity is permuted as a whole. 
As shown in the "Annotator 1" row of Table~\ref{tab_example}, different from the observed two entities with 3 and 1 tokens ("the NIH campus" and "MD"), the left and right positions of the annotated entities in our random model with 3 and 1 tokens ("NIH campus in" and "visited") can be swapped as illustrated in the "Random" column.
With regards to different applications, the random annotation model can be further divided into two sub-models, namely, the overlapping model and the non-overlapping model. 
The overlapping model allows segments to overlap with each other, so each $ST_{i}$ can take any value between $1$ and $n-a_i+1$, whereas the non-overlapping model does not allow segments to overlap, i.e., $ST_{i}\geq ST_{j}+a_{j}$ or $ST_{j}\geq ST_{i}+a_{i}$ for any $i \neq j$. 
Because the overlapping model is much easier to handle, we only focus on the non-overlapping model here.

The problem of estimating chance agreement for annotation evaluation can be described as follows:
	
{\bf \textit{Problem Definition}}. 
Assume there are two independent random annotations, $\Psi1$ for annotator 1 and $\Psi2$ for annotator 2 on the same text of length $n$.
The problem is to estimate the expected similarity $E(Sim(\Psi1,\Psi2))$ based on a random non-overlapping annotation model.

In this paper, we use right index instead of right subscript to represent the index of annotators, for example, $k1$ represents the number of segments annotated by annotator 1, and $k2$ for annotator 2. 
We notice that many agreement measures, regardless of being token level or entity level, can be formulated as segment-wise measures, i.e., \resizebox{.48\textwidth}{!}{$Sim(\psi1,\psi2)=f(\phi_{1,1}(\psi1_{1}, \psi2_{1}), \ldots, \phi_{k1,k2}(\psi1_{k1}, \psi2_{k2}))$}, where $\psi1_{i}=\{st1_{i},a1_{i}\}$ is the ${i}$-th annotated segment for annotator 1 and $\psi2_{j}=\{st2_{j},a2_{j}\}$ is the ${j}$-th one for annotator 2. 
While it is challenging to estimate the chance agreement for a large number of dependent segments together with the random non-overlapping annotation model, the function $f$ is additive for many popular measures. 
This fact allows us to process each segment individually, which greatly simplifies the estimation. 
We call the segment-wise measure with additive function $f$ \textbf{additive measure}.
	
{\bf \textit{Proposition1}}. For the additive similarity measure, the expected chance agreement is \resizebox{.15\textwidth}{!} {$E(Sim(\Psi1,\Psi2))=$} \resizebox{.45\textwidth}{!} {$f(E\phi_{1,1}(\Psi1_{1}, \Psi2_{1})), \ldots, E(\phi_{k1,k2}(\Psi1_{k1}, \Psi2_{k2}))$}. 
	
Note that in the non-overlapping random annotation model, the position of each random annotation segment is dependent on all the other random annotation segments within the same document from the same annotator. 
Since we assume all possible random annotations are equally likely, the problem of estimating the location distribution for each segment is equivalent to counting the number of all possible configurations when we fix the location of the corresponding segment.
	
{\bf \textit{Proposition2}}.
For the non-overlapping random annotation model, the number of all random annotations with the $i$-th segment fixed as:
\begin{equation}
\resizebox{.48\textwidth}{!}{%
  $\begin{aligned}
	& \Pi(ST_i=l) = \pi(l-1,0)\pi(n-l-a+k, k-1)+ \\
	& \sum_{i_1 \neq i} \pi(l-a_{i_1},1) \pi(n-l-a+a_{i_1}+k-1, k-2) + \\
	 & \sum_{i_1 \neq i} \sum_{i_2\neq i}\pi(l-a_{i_1}-a_{i_2}+1,2) 
	 \pi(n-l-a+a_{i_1}+a_{i_2}+k-2, k-3)\\
	 & + \ldots + \pi(l-a+a_i+k-2, k-1)\pi(n-l-a_i+1, 0),\label{eq_1}
  \end{aligned}$%
}
\end{equation}

where $\pi(n,r)=n!/(n-r)!$ is the number of permutations of $n$ things taken $r$ at a time, 
$k$ is the number of segments,
$a_i$ denotes the length of the $i$-th segment and $a=\sum_{i}a_i$ is the total length of annotations. 
Then the corresponding probability is $p(ST_i=l)=\Pi(ST_i=l)/\pi(n-a+k,k)$, for $1\leq l \leq n-a_i+1$.
Here we treat each text segment as a different annotation, regardless of length. 
If we do not need to distinguish among entities of the same length, this formula can also be applied after a simple modification.

However, it is computationally expensive to calculate Equation \ref{eq_1} for all possible random locations of each text segment when the sequence is long. 
To solve this issue, we find that $\Pi(ST_i=l)$ is the same for most locations when the text is of length $n\gg a$.
	
{\bf \textit{Proposition3}}. $ST_i$ is uniformly distributed for $ a-a_i-k+2\leq st_i\leq n-a+k$,
i.e., $\Pi(st_i=l_1)=\Pi(st_i=l_2)$ for $\:\forall\: a-a_i-k+2\leq l_1, l_2\leq n-a+k $.
	


We further observe that it is not necessary to estimate chance agreement in all cases. Intuitively, we expect the chance agreement is small enough to be ignored when annotating sparse information in long texts and find that it is indeed the case.
In most named entity recognition tasks, for example, the average tokens in an annotated sentence is usually large than 20~\citep{roth2004linear}.
	

{\bf \textit{Proposition4}}. When $n\gg a1+a2$, the expected similarity $E(Sim(\Psi1,\Psi2))\to 0$, where $a1$ and $a2$ are the total lengths of all annotated segments for annotator 1 and annotator 2.


For the overlapping model, as the probability of the location of each randomly annotated segment is uniform, we can easily derive its probability distribution.

{\bf \textit{Proposition5}}. 
For the overlapping random annotation model, $p(ST_i=l)=1/(n-a_i+1)$, for $1\leq l \leq n-a_i+1$.

{\bf \textit{Annotation Difficulty Evaluation.}}
Another important application of chance agreement is to define the difficulty of an annotation task from the perspective of agreement by chance. Usually, evaluating the difficulty of annotation tasks is highly subjective and there are no good quantitative indicators. 
We utilize the chance agreement to define the difficulty of annotation tasks as follows:

{\bf \textit{Definition.}} The difficulty level of an annotation task can be defined as $1-E(Sim(\Psi,\Psi))$ if there is a gold standard annotation $\Psi$ or as average similarity of all annotator pairs    $1-\sum_{i,j=1}^{v} E(Sim(\Psi1,\Psi2))/v^2$, where $v$ is the number of annotators. 



\section{Experiments}
To demonstrate the accuracy and effectiveness of our approach, we conducted both simulation and corpus-based experiments\footnote{ All experiments are implemented with MATLAB on a 2017 Mac Pro. The configuration of the Mac Pro is 2.9 GHz Intel Core i7 processor and 16GB 2133 MHz LPDDR3 memory. The evaluation tool and datasets will be released as open-source after the review period.}. We designed the simulation experiments to validate our probability distribution estimation for random sequence annotation. Additionally, by varying the length of text, entity length, and quantity in the simulation experiments, we demonstrated the effectiveness of chance correction, comparing it with Alpha coefficients. Ultimately, we illustrated how our chance estimation impacts the evaluation and ranking of model performance in corpus experiment. Since the estimation of chance agreement for the overlapping model is considerably simpler than for the non-overlapping model, all experiments in this paper are configured with the non-overlapping constraint.

\begin{figure*}[th]
	\centering
	\includegraphics[angle=0,height=0.2\textheight]{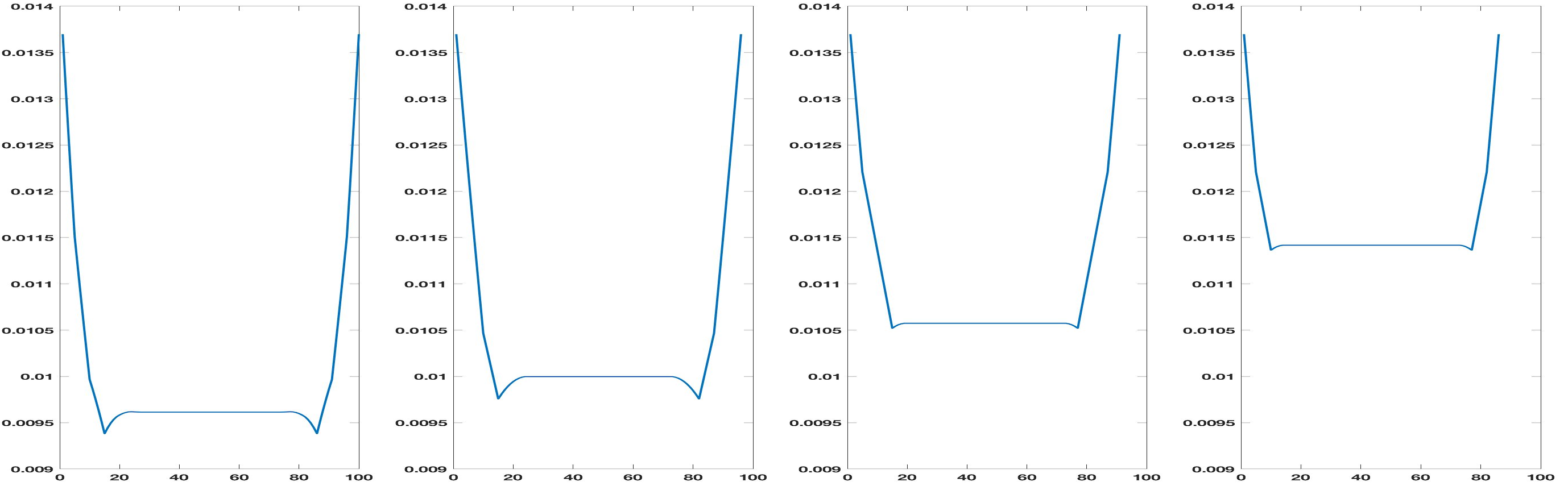}
	\caption{The probability distributions for all possible locations of each  random segment in a length=100 sequence annotated with four segments. The lengths of the four segments are 1, 5, 10, 15, from left to right.}
	\label{fig_text}
    \vspace{0.35cm}
\end{figure*}

Specifically, for the estimation of the probability distribution for random text annotation, we set to label four segments with lengths of 1, 5, 10, and 15 on a sequence of length 100. 
Figure \ref{fig_text} shows the probability distributions of the four segments at all possible locations calculated with the analytical formula in \textit{Proposition 2}. 
The four distributions are approximately distributed as the inverted trapezoids with high ends and flat middle part, which confirms the conclusions of \textit{Proposition 2} and 3.\footnote{The calculation time of the whole process is about 0.01 seconds.}
	 
\begin{table*}[htb]
\begin{center}
\begin{small}
	\begin{tabular}{l|c|c}
	 \toprule
	  & Observed (case A) & Observed (case B)\\
	 \midrule
	 Annotator1 & 0 0 0 \hl{1 1} 0 0 0 \hl{1 1 1} 0 0 0 \hl{1 1 1 1} 0 0 & 0 0 0 \hl{1 1} 0 0 0 \hl{1 1 1} 0 0 0 \hl{1 1 1 1} 0 0 0 0 0 0 0 0 0 0 0 0\\ 
	 Annotator2 & 0 0 \hl{1 1 1} 0 0 0 \hl{1 1 1 1} 0 0 \hl{1 1 1 1 1} 0 & 0 0 \hl{1 1 1} 0 0 0 \hl{1 1 1 1} 0 0 \hl{1 1 1 1 1} 0 0 0 0 0 0 0 0 0 0 0\\
	 \bottomrule
	\end{tabular}
\end{small}
\end{center}
\caption{\label{tab_sim1}
Sequence Annotation Simulation 1.}
\vspace{0.35cm}
\end{table*} 

\begin{table*}[htb!]
\begin{center}
\begin{small}
    \begin{tabular}{l|c|c|c|c|c|c|c|c|c}
    \toprule
    Sim1 & ObsF1 & ChanceF1 & CorrF1 & ObsD & ExpD & Alpha & Obs$\mu$D & Exp$\mu$D & $\mu$Alpha\\
    \midrule
 	 CaseA & 0.8571 & 0.5335 & 0.6938 & 0.0075 & 0.0537 & 0.8602 & 0.15 & 0.5313 & 0.7177 \\ 
 	 CaseB & 0.8571 & 0.3544 & 0.7787 & 0.0033 & 0.0366 & 0.9090 & 0.10 & 0.4704 & 0.7874 \\
 	\bottomrule
 	\end{tabular}
\end{small}
\end{center}
\caption{\label{tab_chance1}
Chance Agreement Estimation for Sequence Annotation Simulation 1.}
\vspace{0.35cm}
\end{table*} 

\begin{table*}[htb!]
\begin{center}
\begin{small}
	\begin{tabular}{l|c|c}
	 \toprule
	  & Observed (case A) & Observed (case B)\\
	 \midrule
	 Annotator1 & 0 0 0 \hl{1 1} 0 0 0 \hl{1 1 1} 0 0 0 \hl{1 1 1 1} 0 0 & 0 0 0 0 0 0 \hl{1 1 1 1 1 1 1 1 1}  0 0 0 0 0\\ 
	 Annotator2 & 0 0 \hl{1 1 1} 0 0 0 \hl{1 1 1 1} 0 0 \hl{1 1 1 1 1} 0 & 0 0 0 0 \hl{1 1 1 1 1 1 1 1 1 1 1 1} 0 0 0 0\\
	 \bottomrule
	\end{tabular}
\end{small}
\end{center}
\caption{\label{tab_sim2}
Sequence Annotation Simulation 2.}
\vspace{0.35cm}
\end{table*} 

\begin{table*}[htb!]
\begin{center}
\begin{small}
	\begin{tabular}{l|c|c|c|c|c|c|c|c|c}
	 \toprule
	  Sim2 & ObsF1 & ChanceF1 & CorrF1 & ObsD & ExpD & Alpha & Obs$\mu$D & Exp$\mu$D & $\mu$Alpha\\
 	 \midrule
 	 CaseA & 0.8571 & 0.5335 & 0.6938 & 0.0075 & 0.0537 & 0.8602 & 0.15 & 0.5313 & 0.7177\\ 
 	 CaseB & 0.8571 & 0.6455 & 0.5970 & 0.0125 & 0.1047 & 0.8806 & 0.15 & 0.5885 & 0.7451\\
 	 \bottomrule
 	\end{tabular}
\end{small}
\end{center}
 \caption{\label{tab_chance2}
 Chance Agreement Estimation for Sequence Annotation Simulation 2.}
\vspace{0.35cm}
 \end{table*} 

\begin{table*}[htb!]
\begin{center}
\begin{small}
	\begin{tabular}{l|c|c}
	 \toprule
	  & Observed (case A) & Observed (case B)\\
	 \midrule
	 Annotator1 & 0 0 0 0 0 0 0 0 \hl{1 1 1} 0 0 0 0 0 0 0 0 0 & 0 0 0 0 0 0 \hl{1 1 1 1 1 1 1 1 1} 0 0 0 0 0\\ 
	 Annotator2 & 0 0 0 0 0 0 0 0 \hl{1 1 1 1} 0 0 0 0 0 0 0 0 & 0 0 0 0 \hl{1 1 1 1 1 1 1 1 1 1 1 1} 0 0 0 0\\
	 \bottomrule
	\end{tabular}
\end{small}
\end{center}
\caption{\label{tab_sim3}
Sequence Annotation Simulation 3.}
\vspace{0.35cm}
\end{table*} 

\begin{table*}[htb!]
\begin{center}
\begin{small}
    \begin{tabular}{l|c|c|c|c|c|c|c|c|c}
    \toprule
	Sim3 & ObsF1 & ChanceF1 & CorrF1 & ObsD & ExpD & Alpha & Obs$\mu$D & Exp$\mu$D & $\mu$Alpha\\
    \midrule
 	CaseA & 0.8571 & 0.1830 & 0.8251 & 0.0025 & 0.0388 & 0.9356 & 0.05 & 0.2996 & 0.8331\\ 
 	CaseB & 0.8571 & 0.6455 & 0.5970 & 0.0125 & 0.1047 & 0.8806 & 0.15 & 0.5885 & 0.7451\\
    \bottomrule
 	\end{tabular}
\end{small}
\end{center}
\caption{\label{tab_chance3}
Chance Agreement Estimation for Sequence Annotation Simulation 3.}
\label{tab_8}
\vspace{0.3cm}
\end{table*} 

\begin{table*}[htb!]
\begin{center}
\begin{small}
	\begin{tabular}{l|c}
	 \toprule
        Gold Standard & \hl{1 1 1} 0 0 \hl{1 1 1} 0 0 \hl{1 1 1} 0 0 \hl{1 1 1} 0 0 \hl{1 1 1} 0 0 \hl{1 1 1 1 1 1 1 1 1 1 1 1 1 1 1 1}\\
        Annotator1 & \hl{1 1 1} 0 0 \hl{1 1 1} 0 0 \hl{1 1 1} 0 0 \hl{1 1 1} 0 0 \hl{1 1 1} 0 0 0 0 0 0 0 0 0 0 0 0 0 0 0 0 0 0\\
        Annotator2 & 0 0 0 0 0 0 0 0 0 0 0 0 0 0 0 0 0 0 0 0 0 0 0 0 0 \hl{1 1 1 1 1 1 1 1 1 1 1 1 1 1 1 1}\\
	 \bottomrule
	\end{tabular}
\end{small}
\end{center}
\caption{\label{tab_sim4}
Sequence Annotation Simulation 4.}
\vspace{0.35cm}
\end{table*} 

\begin{table*}[htb!]
\begin{center}
\begin{small}
	\begin{tabular}{l|c|c|c|c|c|c|c|c|c}
	\toprule
	Sim4 & ObsF1 & ChanceF1 & CorrF1 & ObsD & ExpD & Alpha & Obs$\mu$D & Exp$\mu$D & $\mu$Alpha\\
	\midrule
 	Annotator1 & 0.6522 & 0.5013 & 0.3026 & 0.1523 & 0.2154 & 0.2931 & 0.3902 & 0.5222 & 0.2527\\ 
 	Annotator2 & 0.6808 & 0.5437 & 0.3005 & 0.0268 & 0.2881 & 0.9071 & 0.3659 & 0.5365 & 0.3181\\
 	\bottomrule
	\end{tabular}
\end{small}
\end{center}
\caption{\label{tab_chance4}
Chance Agreement Estimation for Sequence Annotation Simulation 4.}
\vspace{0.3cm}
\end{table*} 

The problem of chance estimation and correction is unique in that, to our knowledge, there is no real benchmark data that can be used to evaluate the performance. Therefore, most classic works in this field use synthetic data to illustrate and evaluate the effect of chance correction, such as \citet{2_komagata2002chance} and \citet{6_artstein2008inter}. Intuitively, we know that the chance agreement is related to the size of the search space, the number of annotated objects, and the lengths of the annotated objects. We design the corresponding comparison experiments by varying these three factors.

\begin{table*}[htb!]
\centering
\begin{center}
\scriptsize
	\begin{tabular}{c|c|c|c|c|c|c|c|c|c|c|c|c|p{0.5cm}}
	 \toprule
	 \multirow{2}{*}{Model} & \multicolumn{4}{c|}{F1-all}  & \multicolumn{4}{c|}{F1-subset1}  & \multicolumn{4}{c|}{F1-subset2} & \multirow{2}{*}{Time} \\
	 \cline{2-13} & Obs & Rank & Cor & Rank & Obs & Rank & Cor & Rank & Obs & Rank & Cor & Rank\\
	 \midrule
	 A & 0.923 & 3 & 0.901 & 3 & 0.919 & 2 & 0.911 & 2 & 0.9369 & \textcolor{red}{3} & 0.9035 & \textcolor{red}{4} & 23\\ 
	 B & 0.905 & 7 & 0.878 & 7 & 0.889 & 7 & 0.878 & 7 & 0.9305 & 6 & 0.8938 & 6 & 23\\
	 C & 0.9072 & 6 & 0.881 & 6 & 0.892 & 6 & 0.881 & 6 & 0.9320 & 5 & 0.8963 & 5 & 23\\
	 D & 0.902 & 8 & 0.874 & 8 & 0.885 & 8 & 0.874 & 8 & 0.9261 & 7 & 0.8878 & 7 & 23\\
	 E & 0.785 & 11 & 0.730 & 11 & 0.731 & 11 & 0.707 & 11 & 0.8537 & 11 & 0.7838 & 11 & 19\\
	 F & 0.846 & 9 & 0.805 & 9 & 0.815 & 9 & 0.798 & 9 & 0.8929 & 9 & 0.8391 & 9 & 18\\
	 G & 0.925 & 2 & 0.904 & 2 & 0.917 & 3 & 0.908 & 3 & 0.9414 & 2 & 0.9103 & 2 & 24\\
	 H & 0.921 & 4 & 0.898 & 4 & 0.913 & 4 & 0.904 & 4 & 0.9368 & \textcolor{red}{4} & 0.9036 & \textcolor{red}{3} & 24\\
	 I & 0.932 & 1 & 0.913 & 1 & 0.922 & 1 & 0.914 & 1 & 0.9500 & 1 & 0.9232 & 1 & 23\\
	 J & 0.9073 & 5 & 0.882 & 5 & 0.903 & 5 & 0.894 & 5 & 0.9240 & 8 & 0.8851 & 8 & 22\\
	 K & 0.802 & 10 & 0.752 & 10 & 0.759 & 10 & 0.737 & 10 & 0.8537 & 10 & 0.7854 & 10 & 16\\ 
	 \bottomrule
	\end{tabular}
\end{center}
\caption{\label{tab_CONLL03}
Chance Agreement Estimation for CoNLL03 Dataset. Obs is short for observed F1 as reported in corresponding real NER model (A-K), Cor is short for corrected F1. Time denotes the running time for chance estimation in seconds.}
\vspace{0.3cm}
\end{table*} 

We design three sets of comparison experiments by varying the length of text (simulation 1), the number (simulation 2) and length (simulation 3) of entities. In case \textit{A} of simulation 1 shown in Table \ref{tab_sim1}, we use 1 or 0 to indicate that each token in the text sequence is labeled or not. 
For the same sequence with 20 tokens, annotator 1 labels 3 entities with lengths of 2, 3, and 4. 
Annotator 2 labels 3 entities with lengths of 3, 4, and 5. 
The annotations of case \textit{B} for two annotators are the same as in case \textit{A}, the only difference is that ten 0s are added after the 20 tokens, that is, neither annotator 1 nor annotator 2 have labeled the extra 10 tokens. 
As reported in Table \ref{tab_chance1}, because F1 score only focuses on the annotated tokens, the observed agreement (F1 score) is the same in both cases. However, since the labeled information in case \textit{B} is relatively sparse, the chance agreement in case \textit{B} is smaller, and the corresponding corrected F1 score is larger which means the agreement is higher. 
In simulation 2, the text length and the total number of annotated tokens remain the same, but the number of annotated entities changes from 3 in case A to 1 in case B. 
In simulation 3, the text length and the number of annotated entities remain the same, whereas the number of annotated tokens in case B is tripled. 
The results in Table \ref{tab_chance1}, \ref{tab_chance2} and \ref{tab_chance3} show that the longer the text, or the more entities, or the shorter the entities, the smaller the chance agreement. This is consistent with our intuition. 

 We also compared our results with two Alpha coefficients, namely Alpha and $\mu$Alpha (see \citealp{krippendorff2016reliability} Equation 2 and Equation 5a for specific formulas). At first glance, Alpha coefficients exhibit a similar trend in simulations 1 and 3, consistent with intuition, while the results in simulation 2 contradict intuition. However, the underlying reasons are different. Our results are derived from chance agreement estimations that align with intuition, whereas the results of Alpha coefficients are influenced by their measurement metrics. For the critical estimation of expected disagreement (ExpD and Exp$\mu$D), it should have an inverse trend with expected agreement (chanceF1) because the more the agreement, the less the disagreement. However, the actual results are the opposite, primarily because Alpha coefficients include agreement for irrelevant segments, which does not align with the needs of most information extraction tasks.
 
The main purpose of chance correction is to use different baselines for different tasks. In addition, chance correction may also change the ranking of model performance for the same task, although this is not common. As shown in the table \ref{tab_sim4}, the gold standard annotation labels six entities with size of 3, 3, 3, 3, 3, 16. The annotator1 labels five 3-token entities correctly but misses the  16-token entity. The annotator2 labels the 16-token entities correctly but misses five 3-token entities. Note that the observed F1 score of annotator1 is lower than that of annotator2. But after the chance correction, the results are opposite (see table \ref{tab_chance4}). Neither of the two Alpha coefficients demonstrated this capability.

To evaluate our model on real data, we estimated the chance agreement of 11 state-of-the-art NER models \citep{liu2021explainaboard} using the CoNLL03 NER dataset \citep{sang2003introduction}. The results are presented in Table \ref{tab_CONLL03}. The CONLL03 testing dataset comprises 3,453 sentences, each annotated with four types of entities: persons (PER), organizations (ORG), locations (LOC), and miscellaneous names (MISC).

We employ a micro-average approach to handle multiple sentences and entity types. This involves separately calculating token-level observed agreement and chance agreement for each sentence and entity type. These token-level observed agreements and chance agreements are then aggregated to compute the overall chance agreement, observed F1 score, and corrected F score. It's important to note that validating chance agreement for real data without ground truth is challenging. However, the F1 scores demonstrate a noticeable widening of the range after chance correction.

Furthermore, we partition the entire 3,453 sentences of the CoNLL03 data into two roughly equivalent subsets based on the chance agreement level for each sentence. Subset1 consists of sentences with a chance agreement level greater than 0.825 (equivalent to difficulty level less than or equal to 0.175), while subset2 includes sentences with a chance agreement level less than or equal to 0.825 (equivalent to difficulty level greater than or equal to 0.175). The results indicate significant changes in the performance ranking of the 11 NER models across different datasets. Additionally, the performance ranking of all 11 models on subset2 also exhibits slight variations before and after chance correction.

\section{Conclusion and Discussion}
In this paper, we propose a novel sequence random annotation model that takes into account the different annotation styles of annotators and the characteristics of sequence annotations. 
For complex cases where labeled objects are required to be disjoint, we investigate the corresponding distribution characteristic and remove redundant calculations. 
We also derive an analytical formula to calculate the exact distribution. 
Our focus in this work is how to establish a general framework and corresponding fast algorithm for calculating similarity by chance in complex text annotations. 
The framework and method proposed in this paper are applicable to all additive similarity measures. Moreover, our approach can extend to nested spans by iteratively applying the same method layer by layer, ensuring compliance with the nested structure.

\section{Limitations}
Since chance estimation for sequence annotation is an open problem, there is very limited similar work to provide as a baseline for direct comparison. In addition, chance estimation lacks benchmark data with ground truth, although we have applied it to real data in order to demonstrate its utility. The current analysis of its effectiveness is mainly based on simulated data and whether it is consistent with human intuition. We expect that this work will stimulate more related work and benchmark data creation. The chance estimation in this paper focuses on the comparison between two annotators, and we plan to extend it to team-wise agreement for more than two annotators or systems.
\section{Ethics Statement}
The use of data on this project strictly adhered to ethical standards required by the National Institute of Health (NIH).

In addition to upholding ethical principles in conducting this work, we believe this work contributes to professional standards for rigor in the field.  In particular, we expect that this paper will facilitate fair comparison of various annotation tasks or systems and reduce random chance agreement caused by different annotation styles and metrics. Chance agreement can also be used as a quantitative aid to measure the difficulty of annotation task. This provides a new perspective for evaluating different annotation tasks.

\section{Acknowledgements}
This study was supported by the Social Security Administration-
National Institutes of Health Interagency Agreements and by the
National Institutes of Health Intramural Research program.
\bibliography{anthology,custom}
\bibliographystyle{acl_natbib}

\newpage
\section{Appendix}

	{\bf \textit{Proposition1}} For the additive similarity measure, the expected chance agreement is $E(Sim(\Psi1,\Psi2))=f(E\phi_{1,1}(\Psi1_{1}, \Psi2_{1})), \ldots, E(\phi_{k1,k2}(\Psi1_{k1}, \Psi2_{k2})))$. 
	
	{\bf \textit{Proof}}.

    Since the function $f$ is additive, the order of the function $f$ and expectation can be interchanged. 
    We have $E(Sim(\Psi1,\Psi2))=E(f(\phi_{1,1}(\Psi1_{1}, \Psi2_{1}), \ldots, \phi_{k1,k2}(\Psi1_{k1}, \Psi2_{k2})))=f(E(\phi_{1,1}(\Psi1_{1}, \Psi2_{1})), \ldots, E(\phi_{k1,k2}(\Psi1_{k1}, \Psi2_{k2})))$.
	
    Originally, to estimate the expectation of similarity by chance, we need to sum up the similarity in a high-dimensional space of all possible random annotations, i.e., $E(Sim(\Psi1,\Psi2))=\sum_{\Psi1_{1}} \ldots\sum_{\Psi1_{k1}}$ $\sum_{\Psi2_{1}} \ldots\sum_{\Psi2_{k2}}f(.)\times p(\Psi1_{1}=\psi1_1, \ldots, \Psi2_{k2}=\psi2_{k2})$. 
    Now we can simplify it to multiple low-dimensional summations, such as $E(\phi_{i,j}(\Psi1_{i}, \Psi2_{j}))$,  under the condition of additive measure.
	
    Note that in the non-overlapping random annotation model, the position of each random annotation segment is dependent on all the other random annotation segments within the same document from the same annotator. 
    Since we assume all possible random annotations are equally likely, the problem of estimating the location distribution for each segment is equivalent to count the number of all possible configurations when we fix the location of the corresponding segment.

 {\bf \textit{Proposition2}}
For the non-overlapping random annotation model, the number of all random annotations with the $i$-th segment fixed as:
\begin{equation}
\resizebox{.48\textwidth}{!}{%
  $\begin{aligned}
	& \Pi(ST_i=l) = \pi(l-1,0)\pi(n-l-a+k, k-1)+ \\
	& \sum_{i_1 \neq i} \pi(l-a_{i_1},1) \pi(n-l-a+a_{i_1}+k-1, k-2) + \\
	 & \sum_{i_1 \neq i} \sum_{i_2\neq i}\pi(l-a_{i_1}-a_{i_2}+1,2) 
	 \pi(n-l-a+a_{i_1}+a_{i_2}+k-2, k-3)\\
	 & + \ldots + \pi(l-a+a_i+k-2, k-1)\pi(n-l-a_i+1, 0),\label{eq_1}
  \end{aligned}$%
}
\end{equation}

where $\pi(n,r)=n!/(n-r)!$ is the number of permutations of $n$ things taken $r$ at a time, 
$k$ is the number of segments,
$a_i$ denotes the length of the $i$-th segment and $a=\sum_{i}a_i$ is the total length of annotations. 
Then the corresponding probability is $p(ST_i=l)=\Pi(ST_i=l)/\pi(n-a+k,k)$, for $1\leq l \leq n-a_i+1$.
Here we treat each text segment as a different annotation, regardless of whether they have the same length. 
If we do not need to distinguish among entities of the same length, this formula can also be applied after a simple modification.

{\bf \textit{Proof sketch}}. 
We can divide all possible random annotations with $ST_i=l$ into $k$ disjoint sets with $m$ annotation segments located on the left of the specified $i$-th segment $\psi_i$ and the remaining $k-m-1$ segments on the right side. 
The cardinality of each set with selected left $m$ annotation segments (which then determines the segments on the right )
is the number of all possible annotations on the left $l-1$ times the number for $n-l-a_i$ of tokens on the right side. 
	
If we fix the order of $m$ selected random annotation segments $\psi_{i_1}$, ..., $\psi_{i_m}$,  the random annotation of the left $l-1$ tokens is equivalent to distribute $l-1-\sum_{j=1}^{m}a_{i_j}$
objects into $m+1$ spaces, before the first annotation segment, between adjacent segments, and after the last one.  This is a well studied problem (integer weak composition into a fixed number of parts) with  $(l-1-\sum_{j=1}^{m}a_{i_j}+m)!/(l-1-\sum_{j=1}^{m}a_{i_j})!/m!$ possible configurations. 
Since we treat all annotation segments as different ones, there are $m!$ permutations for the left $m$ segments and $(k-m-1)!$ for the right $k-m-1$ ones, and the cardinality of each set is $\pi(l-\sum_{j=1}^{m}a_{i_j}+m-1,m)\times \pi(n-l-a+\sum_{j=1}^{m}a_{i_j}+k-m, k-m-1)$.
Based on the above derivation, the number of all possible configurations when we fix the location of a segment can be expressed by Equation \ref{eq_1}.

However, it is computationally expensive to calculate Equation \ref{eq_1} for all possible random locations of each text segment when the sequence is very long. 
To solve this issue, we find that $\Pi(ST_i=l)$ is the same for most locations when the text is of length $n\gg a$. Please note that the effectiveness of {\bf Proposition3} is not related to the length of the sentence. It's just that the longer the sentence, the more computation Proposition 3 can reduce. For short sentences, the computational cost itself is not significant.
	
	\noindent{\bf Proposition3}. $ST_i$ is uniformly distributed for $ a-a_i-k+2\leq st_i\leq n-a+k$, i.e., $\Pi(st_i=l_1)=\Pi(st_i=l_2) \:\forall\: a-a_i-k+2\leq l_1, l_2\leq n-a+k $ .
	
	It is clear that proposition  3 and proposition 3*  are equivalent.
	
	\noindent{\bf Proposition3*}. $\Pi(st_i=l)=\Pi(st_i=l+1) \:\forall\: a-a_i-k+2\leq l\leq n-a+k-1 $ .
	
	{\bf Proof sketch}. 
	\noindent Use mathematical induction
	
	\noindent Initial step: when $k=1$, $\Pi(st_1=l)=1$ and $p(st_1=l)=1/(n-a_1+1)$, for $1\leq l\leq n-a_1+1$. So the proposition 3* is true at $k=1$.
	
	\noindent Inductive step: assume the proposition 3* holds for $k=r$. When $k=r+1$, we partition all possible configurations with $st_i=l$ into $r+1$ disjoint scenarios: the $r$ scenarios with $st_j=l+a_i$ for all $j\neq i$ and the rest,  i.e., the scenarios with a different annotation segment next to $\psi_i$ from right side or none annotation segment next to $\psi_i$ from right side. So $\Pi(st_i=l)=\sum_{j\neq i}\Pi(st_i=l \: \&\:  st_j=l+a_i) + \Pi(st_i=l \: \& \: st_j\neq l+a_i, \forall j \neq i )$. 
	
	We also partition all possible configurations with $st_i=l+1$ into $r+1$ disjoint scenarios: the $r$ scenarios with $st_j=l+1-a_j$ for all $j\neq i$ and the rest, i.e., the scenarios with a different annotation segment next to $\psi_i$ from left side or none annotation segment next to $\psi_i$ from left side. Similarly, $\Pi(st_i=l+1)=\sum_{j\neq i}\Pi(st_i=l+1 \: \& \: st_j=l+1-a_j) + \Pi(st_i=l +1 \:\& \: st_j\neq l+1-a_j, \forall j \neq i )$.
	
	Since there is a bijection between the scenario of $st_i=l \:\&\:  st_j\neq l+a_i, \forall j \neq i $ and the one of $st_i=l+1 \:\:\&\:\:  st_j\neq l+1-a_j, \forall j \neq i $ by identity mapping except the annotation segment $\psi_i$ and the un-annotated token next to it with indices from $l$ to $l+a_i$, $\Pi(st_i=l \:\& \: st_j\neq l+a_i, \forall j \neq i )=\Pi(st_i=l +1  \&  st_j\neq l+1-a_j, \forall j \neq i )$. For the pair of scenarios $st_i=l \:\&\:  st_j=l+a_i$ and $st_i=l+1 \: \& \: st_j=l+1-a_j$, they can be convert to scenarios $st_i^*=l \: \& \: a_i^*=a_i+a_j$ and $st_i^*=l+1-a_j \: \& \: a_i^*=a_i+a_j$ by merging $\psi_i$ and $\psi_j$. Based on the assumption that the proposition 3* holds at $k=r$, their cardinalities should be equal since there is only $r$ segments after the combination and $a-(a_i+a_j)-(k-1)+2\leq l,l+1-a_j\leq n-a+(k-1) $. Therefore, $\Pi(st_i=l \: \&\:  st_j=l+a_i)=\Pi(st_i=l+1 \: \& \: st_j=l+1-a_j)$ and the proposition 3* holds for $k=r+1$.
	
	It is a tight bound since we have to satisfy the condition of $0\leq l-\sum_{j=1}^{m}a_{i_j}+m-1$ and $0\leq n-l-a+\sum_{j=1}^{m}a_{i_j}+k-m$ for all $0\leq m \leq k-1$ and $i_j \neq i$. This is the same as $a-a_i-k+2\leq l\leq n-a+k$.	
	
	\begin{figure}[hbt!]
	\centering
	\includegraphics[angle=0,height=0.15\textheight]{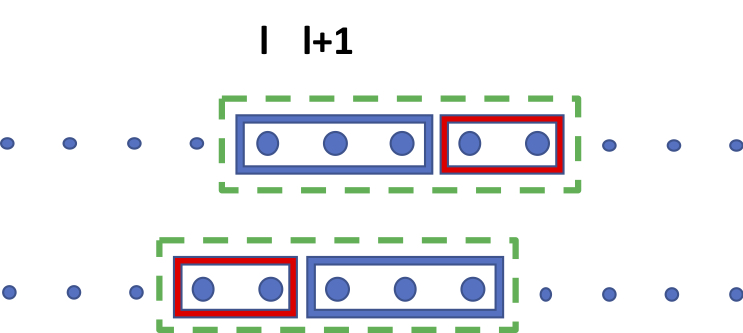}
	\caption{Convert the case of $k=r+1$ to the case of $k=r$ by merging two adjacent text segments $\psi_i$ and $\psi_j$, the blue box represents the segment $\psi_i$ , and the red box represents the adjacent segment $\psi_j$.}
	\end{figure}

    {\bf \textit{Proposition4}}. The expected similarity $E(Sim(\Psi1,\Psi2))\to 0$ when $n\gg a1+a2$, where $a1$ and $a2$ are the total lengths of all annotated segments for annotator 1 and annotator 2. 

{\bf \textit{Proof sketch}}. 
According to the proof process of \textit{Proposition} 2, we know the number of all possible random annotations of $k$ segments with total length $a$ for a text with $n$ tokens is $\pi(n-a+k,k)$. 
Thus, the total number of comparisons between random annotations from annotator 1 and annotator 2 is $\pi(n-a1+k1,k1)\times \pi(n-a2+k2,k2)$ under the independent annotation assumption. 
It is straight forward that the segment-wise agreement $\phi_{i_1,i_2}(\psi1_{i_1}, \psi2_{i_2})$ is zero if there is no overlap between the $i_1$-th text segment annotated by annotator 1 and the $i_2$-th text segment annotated by annotator 2. 
The agreement between two annotators is zero if there is no overlap among all $k1+k2$ annotated text segments. 
The situation is equivalent to combining the annotation results of the two annotators and requiring no overlap among all $k1+k2$ text segments in the same text. 
The total number of such possible annotations is $\pi(n-a1-a2+k1+k2,k1+k2)$. 
Therefore, the probability of zero chance agreement $p(Sim(\Psi1,\Psi2))=0)=\pi(n-a1-a2+k1+k2,k1+k2)/\pi(n-a1+k1,k1)/\pi(n-a2+k2,k2)=(n-a1-a2+k1+k2)\times \ldots (n-a1-a2+1)/((n-a1+k1)\times \ldots (n-a1+1)\times (n-a2+k2)\times \ldots (n-a2+1))\to 1$ because both numerator and denominator are to the $(k1+k2)$-th power of $n$ and $n \gg a1+a2 \geq k1+k2$. 
Thus, we have $E(Sim(\Psi1,\Psi2))\to 0$ when $n \gg a1+a2$.

{\bf \textit{Proposition5}}. 
For the overlapping random annotation model, $p(ST_i=l)=1/(n-a_i+1)$, for $1\leq l \leq n-a_i+1$.
	
{\bf \textit{Proof sketch}}. This conclusion is straight forward because a random text segment annotation with length $a_i$ can be placed at any feasible locations with equal probability without the non-overlapping constraint.
    
	\noindent{\bf Computational complexity for random text annotation}. The computational cost of calculating the probability distribution of the location of $k$ random annotated  text segments is bounded by $((k-1)\times a-k^2+2k)\times 2^k \times (k-1)$ multiplications and $((k-1)\times a-k^2+2k)\times (2^k-1)$ additions.
	
	In order to calculate the probability distributions for random text annotation, according to the proposition 2 and the proposition  3, we could calculate the probability of $ a-a_i-k+2$ possible positions for each random annotated text segment with formula 1.  And the analytical formula is a summation of $2^k$ terms, and each term is equivalent to $k-1$ multiplications, so the computational complexity is bounded by  $\sum_{i=1}^{k}(a-a_i-k+2)\times 2^k \times (k-1)=((k-1)\times a-k^2+2k)\times 2^k \times (k-1)$ multiplications and $\sum_{i=1}^{k}(a-a_i-k+2)\times (2^k-1) =((k-1)\times a-k^2+2k)\times (2^k-1)$ additions. Since the formula 1 is a subset convolution, It may be possible to speed up this calculation with the fast subset convolution algorithm.
	
	According to the above computational complexity analysis, we know that the probability distribution of the location of each random annotated segment can be calculated efficiently using the formula 1 when the number of text segments $k$ is small. But with the increase of $k$, the computational cost will increase rapidly. Fortunately, when the text  sequence is long enough and the annotated information is sparse, we can use the uniform distribution to approximate the distribution. 
	
	\noindent{\bf Uniform approximation}. The probability distribution of the location of a random annotated text segment can be approximated by uniform distribution with $p(st_i=l)=1/(n-a_i+1)$, for $1\leq l\leq n-a_i+1$ if $(n-a+k)/(n-a_i+1)>\alpha$, where $\alpha$ is a preset threshold which is close to 1 and less than 1, for example $\alpha=0.99$ .
	
	We observe that the probability distribution of the location of a random annotated text segment is approximately inverted trapezoid distributed with highest probabilities at both ends. And the majority of the whole distribution is flat when $n>>a$. It is straight forward to calculate the $p(st_i=1)=\pi(n-a+k-1,k-1)/\pi(n-a+k,k)=1/(n-a+k)$. So the distribution could be approximate with uniform distribution if the highest probability $1/(n-a+k)$ is close to the uniform probability $1/(n-a_i+1)$, i.e., $(n-a+k)/(n-a_i+1)$ is close to 1 if $n>>a$.

 {\bf \textit{CoNLL03 NER dataset and system outputs}}.
 To evaluate our model in real data, we estimate the chance agreement of 11 state-of-the-art NER models on CoNLL03 NER dataset, the results are shown in Table \ref{tab_CONLL03}. CoNLL-2003 is a named entity recognition dataset that is released as a part of CoNLL-2003 shared task: language-independent named entity recognition. This corpus consists of Reuters news stories between August 1996 and August 1997. There are four types of annotated entities: persons (PER), organizations (ORG), locations (LOC) and miscellaneous names (MISC). We downloaded 15 system outputs for the English test set from the Explained Board website after approval. Since 4 system outputs use different sentence segmentation, we limit our comparison to 11 system outputs that use the same sentence segmentation. The test set consists of 231 articles that include 3453 sentences.

\end{document}